# FACE RECOGNITION IN A TRANSFORMED DOMAIN[1]

**Marcos Faúndez-Zanuy**
Dep. de Telecomunicaciones y Arquitectura de computadores
Escuela Universitaria Politécnica de Mataró, Adscrita a la UPC
Avda. Puig i Cadafalch 101-111, 08303 Mataró (Barcelona), SPAIN
e-mail: faundez@eupmt.es

## ABSTRACT

This Paper proposes the use of a Discrete Cosine Transform (DCT) instead of the eigenfaces method (Karhunen-Loeve Transform) for biometric identification based on frontal face images. Experimental results show better recognition accuracies and reduced computational burden.
This paper includes results with different classifiers and a combination of them.

## 1. INTRODUCTION

When dealing with images, a model of production can not be supposed, but the light-intensity function at each pixel of the image can be decomposed into two components [1, p.30]:
1. Illumination (i(x,y)) or the amount of source light incident on the scene.
2. Reflectance (r(x,y)) or the amount of light reflected by the objects in the scene.

Then, the bidimensional light-intensity function, represented by f(x,y), can be obtained as: $f(x,y) = i(x,y)r(x,y)$, where the nature of i(x,y) is determined by the light source, and r(x,y) is determined by the characteristics of the objects in a scene. Obviously, for image recognition the useful component is only the last one, because the illumination is irrelevant and its effect over the recognition rates is harmful.

In speech and speaker recognition applications, better results are achieved using a parameterization of the signals in a transformed domain. Usually, it is used the logarithm of the Fourier transform, known as Cepstrum. For speech signals, a model of production can be assumed. Mainly it consists on two parts: the excitation signal, and the vocal tract filter. For this model, the Cepstrum achieves the deconvolution between these two kinds of information, being the vocal tract information the most important for speech and speaker recognition.

Illumination and reflectance components of images can be separated using a procedure analogous to the speech processing deconvolution Cepstrum. In the image case, the components to be separated are multiplied rather than convoluted, so the logarithm function must be performed in the first step, and the next operation must be a transform function, because both components have different spectral contents. That is, the illumination is a low frequency function, so in a transformed domain it would be concentrated in different coefficients than the reflectance component. Thus, it will be easy to apply a filter for removing the effect of the illumination.

$$FFT\{\ln(f(x,y))\} = FFT\{\ln(i(x,y))\} + FFT\{\ln(r(x,y))\}$$

For image enhancement, the filtered image must be inverse transformed, and the exponential function must be applied for compensating the effect of the logarithm function. For image recognition, it can be performed in the transformed domain, because there are additional advantages. Mainly, the energy compactation introduced by the transformation lets a dimensionality reduction, and thus a lower computational burden. The classical eigenfaces method [2] uses a data-dependent transformation (Karhunen Loeve Transform), but we propose the use of de Discrete Fourier Transform (DFT) or the Discrete Cosine Transform (DCT), because they are faster and in our simulation results outperform the eigenfaces method. It must be taken into account that in image coding applications the DCT is preferred, rather than KLT (for example JPEG or MPEG use the DCT). A comparison between these transforms reveals [3, pp.179]:

- KL: although it is optimal in many ways, it does not have a fast algorithm. It is useful in performance evaluation and for finding performance bounds. It is useful for small size vectors (that is not the case in face recognition). It has the best energy compactation in the mean square sense over an ensemble.
- DCT: It is a fast transform that requires real operations and it is a near optimal substitute for the KL transform of highly correlated images. It has excellent energy compactation for images.
- DFT: Fast transform, most useful in signal processing (de/) convolution. Requires complex arithmetic, and has very good energy compactation for images.

On the other hand it must be taken into account that usually the KLT is computed using an approximate solution, due to the high dimensional size of the covariance matrix [2]. This implies that some reduction on the upper performance bound exists with respect to the exact formulation of the KL transform. On the other hand in face recognition the number of sample vectors is usually much smaller than the vector dimension, so the rank of the sample covariance matrix is smaller or equal to the number of available samples (this is the small sample size problem, [4, p.39]), so the number of features than can be extracted with the KL is limited.

---

[1] This work has been supported by the CICYT TIC2000-1669-C04-02

**The relevance of vector dimension reduction**

Usually a pattern recognition system consist on two main blocks: feature extraction and classifier. Figure 1 summarizes this scheme. On the other hand, two main approaches for face recognition exist:
a) Statistical approaches consider the image as a high dimensional vector, where each pixel is mapped to a component of a vector. Due to the high dimensionality of the vectors some vector dimension reduction algorithm must be used. Typically the Karhunen Loeve transform is applied [5].
b) Geometry-feature-based methods try to identify the position and relationship between face parts, such as eyes, nose, mouth, etc., and the extracted parameters are measures of textures, shapes, sizes, etc. of these regions.

While great efforts have been done on feature extraction for geometrical methods, in statistical approaches few efforts have been done to improve the feature extraction, and almost all the systems are based on the Karhunen-Loeve dimensionality reduction method.

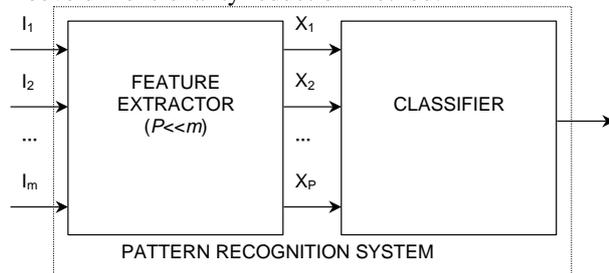

*Figure 1. General pattern recognition system*

In this paper, we mainly focus on the study of the feature extraction for the face recognition using statistical approaches. On the final part we also study the use of neural network classifiers and a committee machine of different classifiers with the goal to improve the recognition accuracy.

The relevance of feature extraction is twofold:
- It achieves a reduction on the number of data that must be processed, model sizes, etc., with the consequent reduction on computational burden.
- The transformation of the original data into a new feature space can let an easier discrimination between classes (faces).

On the other hand it is not trivial to decide which are the most suitable features and/ or what is the optimal dimension. Even the best set of features for a given classifiers can be suboptimal for a different classifier. For instance, the optimal set for a linear classifier can not be optimal for a neural net classifier [4, pp.441].

In order to illustrate the relevance of vector dimensions, we show a toy problem that consist on the classification of two different classes (rhombus and circles) using two features per class (P=2, two dimensional feature space). If we try to reduce the vector dimension with the simplest procedure, we just need to remove the first or second component. For the classes distribution shown in figure 2 it is clear that none of the components alone lets to classify correctly(there is a strong overlap over one single axis), but both together can easily achieve this purpose.

On the other hand the distribution shown in figure 3 shows that the first component has no discriminative power, while there is no overlap on the second dimension. In this case, if an Euclidean distance based classifier is used, better results will be obtained removing the first component. This is because the distance between an input vector $\vec{x} = (x_1, x_2)$ and any of the classes representatives (for instance the mean of the distribution of each class) $\vec{y} = (y_1, y_2)$ would be:

$$d^2(\vec{x}, \vec{y}) = (x_1 - y_1)^2 + (x_2 - y_2)^2 \cong (x_1 - y_1)^2,$$

because $(x_1 - y_1)^2 >> (x_2 - y_2)^2$. Thus the distance measure is masked by the first component, because it has higher variance that the second component.

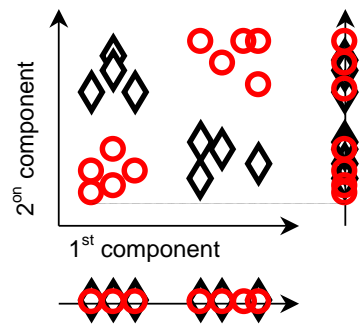

*Figure 2. Example of two class separation using two dimensional vectors*

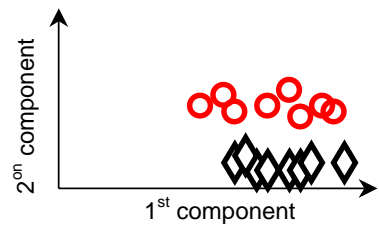

*Figure 3. Example of two class separation using two dimensional vectors*

Mathematically the discriminative ability for a two class problem can be estimated by the distance between means normalized by the variance [6, p.91]:

$$D_{1ij} = \frac{|m_{1i} - m_{1j}|}{\sqrt{\sigma_{1i}^2 + \sigma_{1j}^2}}, \quad D_{2ij} = \frac{|m_{2i} - m_{2j}|}{\sqrt{\sigma_{2i}^2 + \sigma_{2j}^2}}$$

Where:
- The two clases are $i$ (for instance circles) and $j$ (rhombus)
- The features that we are studying are 1 and 2 (first and second component).
- The mean and variance of the classes can be obtained with :

$$m_{1i} = \frac{1}{N_i}\sum_{k=1}^{N_i} x_{1ki}, \quad \sigma_{1i}^2 = \frac{1}{N_i}\sum_{k=1}^{N_i}(x_{1ki} - m_{1i})^2$$

$$m_{2i} = \frac{1}{N_i}\sum_{k=1}^{N_i} x_{2ki}, \quad \sigma_{2i}^2 = \frac{1}{N_i}\sum_{k=1}^{N_i}(x_{2ki} - m_{2i})^2$$

Analogous formulation would be used for the class $j$ using vectors that belong to this class for the following results: $m_{1j}, m_{2j}, \sigma_{1j}^2, \sigma_{2j}^2$. For the example shown in figure 3 the results would be $D_{2ij} \gg D_{1ij}$, because $|m_{2i} - m_{2j}| > |m_{1i} - m_{1j}|$, $\sqrt{\sigma_{2i}^2 + \sigma_{2j}^2} < \sqrt{\sigma_{1i}^2 + \sigma_{1j}^2}$. The most interesting features should be selected taking into account the discriminative power. This features are known as the Most Discriminative Features (MDF). However, they use to be selected the most expressive features (MEF) [7, pp.71], also known as eigenfaces, or at least they are used as a first step before the computation of the MDF.

## 2. THE EIGENFACE APPROACH

Turk and Pentland [2], propose an eigenface system which projects face images onto a feature space that spans the significant variations among known face images using the Karhunen-Loéve Transform. It is an orthogonal lineal transform of the signal that concentrates the maximum information of the signal with the minimum number of parameters using the minimum square error (MSE). The significant features are known as eigenfaces, because they are the eigenvectors (principal components) of the set of images. The projection operation characterizes an individual face by a weighted sum of the eigenface features, and so to recognize a particular face it is only necessary to compare these weights to those of known individuals.

**Formulation for Eigenfaces computation**
The computations assume a training sequence of M images $\{I_1, I_2, \ldots, I_M\}$ of size NxN (Altough the images must not be squared we will assume that they are squared without lost of generality). This set of images belongs to a number smaller or equal to M persons (if there is one or more face images for each person).

The eigenfaces method [2] can be summarized in the following steps:
- The images are re-arranged into a one-dimensional vector of $1xN^2$ components.
- To compute the average face:

$$\Psi = \frac{1}{M}\sum_{n=1}^{M} I_n$$

- The average face is subtracted to all the vectors:
$\Phi_i = I_i - \Psi$
- To compute the covariance matrix, with the following expression,

$$C = \frac{1}{M}\sum_{n=1}^{M} \Phi_n \Phi_n^T$$

  where T means the transpose of the vector.
- This matrix can be expressed as:

$$C = AA^T$$

  where the matrix $A = [\Phi_1 \Phi_2 \ldots \Phi_M]$.
- This matrix is diagonalized and the u eigenvectors ($u \ll N^2$) associated to the u greatest eigenvalues are selected. Each of these eigenvectors is named eigenface. Thus, it is possible to obtain a more compacted representation. The geometric interpretation is that each face is approximated by a linear combination of eigenvectors of the selected subspace. The matrix C has a dimension $N^2$ x $N^2$ elements. Thus, for a N=92x112 pixels/image the diagonalization of the matrix implies a high computational burden and memory requirements. For this reason, usually the computations are done with the matrix $A^TA$, which is smaller. In this case, using the eigenvectors $\vec{v}_l = [v_{l1}, v_{l2}, \ldots, v_{lM}]$ can be derived an approximated expression for the eigenvectors $\vec{u}_l = [u_{l1}, u_{l2}, \ldots, u_{lN^2}]$:

$$\vec{u}_l = \sum_{k=1}^{M} \vec{v}_{lk}\Phi_k \qquad l = 1,\ldots,M$$

- For the testing process, the coordinates of the test face ($\Omega$) must be compared against the coordinates of the modeled persons ($\Omega_K$). This can be done projecting the test face into the subspace computed previously. It is not needed to use the M eigenvectors obtained. It is possible to use the M'<M more significatives. The projection of the input face (I) over the M' directions of the vectorial space can be implemented in the following way: $\omega_k = \vec{u}_k^T(I - \Psi)$ where the coordinates $\omega$ form the vector $\Omega^T = [\omega_1, \omega_2, \ldots, \omega_{M'}]$ that parameterizes the input face.

**Face Recognition**
Recognition is performed by finding the training face that minimizes the face distance with respect to the input test face. In other terms, the identification of the test image is done locating the database entry, whose weights are closest (in euclidean distance) to the weights of the face.

## 3. THE DFT AND FFT APPROACH

The bidimensional Discrete Fourier Transform is defined as $F(p,q) = \sum_{m=0}^{M-1}\sum_{n=0}^{N-1} A(m,n) e^{-j\frac{2\pi}{M}pm} e^{-j\frac{2\pi}{N}qn}$

The Discrete Cosine Transform (DCT) is closely related to the discrete Fourier transform. It is a separable, linear transformation; that is, the two-dimensional transform is equivalent to a one-dimensional DCT performed along a single dimension followed by a one-dimensional DCT in the other one. The definition of the two-dimensional DCT for an input image *A* and transformed image *B* is the following:

$$B_{pq} = \alpha_p \alpha_q \left\{ \sum_{m=0}^{M-1}\sum_{n=0}^{N-1} A_{mn} \cos\left(\frac{\pi(2m+1)p}{2M}\right) \times \right.$$

$$\left. \times \cos\left(\frac{\pi(2m+1)q}{2N}\right) \right\}$$

where:
- $0 \le p \le M-1$ and $0 \le q \le N-1$
- $\alpha_p = \begin{cases} \frac{1}{\sqrt{M}}, & p=0 \\ \sqrt{\frac{2}{M}}, & 1 \le p \le M-1 \end{cases}$,
- $\alpha_q = \begin{cases} \frac{1}{\sqrt{N}}, & q=0 \\ \sqrt{\frac{2}{N}}, & 1 \le q \le N-1 \end{cases}$
- *M* and *N* are the row and column size of A, respectively.

The application of the DCT to an image (real data), produces a real result. The DCT tends to concentrate information, making it useful for image compression applications, dimensionality reduction, etc.

Figure 4 shows the first image of the ORL [8] database and the corresponding DCT. It's easy to observe that most of the energy is concentrated around the origin (low frequency components, located on the upper left corner).

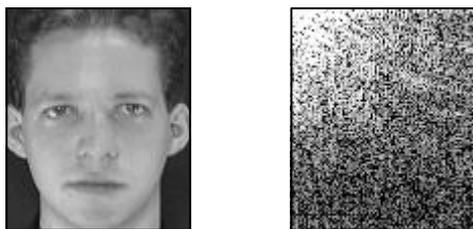

Figure 4: Example of face image, and DCT of this image

One advantage of the DFT and the DCT is that it is possible to apply filters on the transform domain. We can define a zonal mask as the array $m(f_1,f_2) = \begin{cases} 1, & f_1,f_2 \in I_t \\ 0, & otherwise \end{cases}$, and multiply the transformed image by the zonal mask, that takes the unity value in the zone to be retained and zero on the zone to be discarded. In image coding it is usual to define the zonal mask taking into account the transformed coefficients with largest variances. Then the zonal mask is applied to the transformed image (or blocks of the image) and only the nonzero elements are encoded. In our case we will not take into account the variances of the transformed coefficients and we will just define the zonal mask in the following easy ways:
a) Rectangular mask: it will be a square containing *N'*x*N'* pixels
b) Sectorial mask: it will be a sector of 90º of a *r* radius circle.

Figure 5 shows one example of each situation.

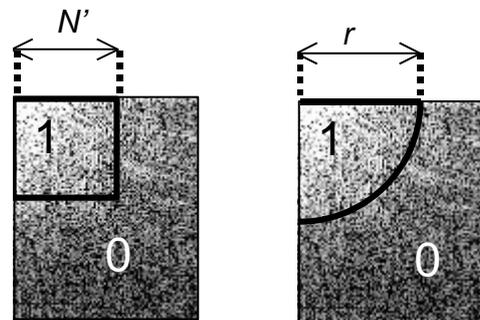

Figure 5: Example of rectangular and sectorial masks.

This definition lets to easily obtain the coefficients. The dimension of the resulting vector is *N'*x*N'* for the rectangular mask and for the sectorial mask the number of pixels that satisfy the following condition:

$\vee f_1, f_2, \text{ if } \sqrt{(f_1-c_1)^2 + (f_2-c_2)^2} < radius$
$\text{then } m(f_1,f_2)=1; \text{ else } m(f_1,f_2)=0$

where the coordinates of the center are the frequency origin ($c_1 = c_2 = 0$).

Obviously other kind of filters and zonal mask definitions are also possible. It is specially interesting the consideration of a band pass filter that removes the very low frequency components. This kind of filter can remove the illumination effect, relying on the principle that was described on the introductory section of this paper. This is a real situation on face recognition applications, where the illumination conditions can vary significatively. On the other hand, if the images are acquired in a good controlled environment (such as the ORL database), this low frequency removal can produce a small decrease on the identification rates.

It is interesting to observe that in image coding applications the image is split into blocks of smaller size, and the selected transformed coefficients of each block are encoded and used for the reconstruction of the decoded image. In face recognition all the operations are performed over the whole image (it is not split into blocks) and all the computations are done in the transformed domain. Thus, it is not necessary to perform any inverse transform. On the other hand in image coding the goal is to reduce the amount of bits without sacrificing appreciably the quality of the reconstructed image, and in image recognition it doesn't matter how much bits are necessary. The goal is to reduce the dimensionality of the vectors in order to simplify the complexity of the classifier and to improve the recognition accuracy. Anyway, figure 6 illustrates the dimensionality reduction that can be achieved through the energy compactation due to the DCT transform. On the left it is the inversed transformed image obtained from a rectangular zonal masking of $N'$=32 pixels. Considering the whole image as a single vector, this implies a dimensionality reduction of 10 times.

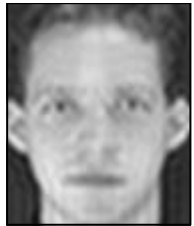

Figure 6 Decoded image after dimensionality reduction by 10.

It must be taken into account that the product between the transformed image and the mask is equivalent to a convolution in the frequency domain, so better results could be obtained with a different kind of window with softer transitions (for instance hamming window instead of rectangular window), but this is irrelevant to this work, because we are just interested on energy compactation and dimensionality reduction.

## 4. EXPERIMENTAL RESULTS

This section evaluates the results obtained using the DCT and compares them with the classical eigenface method. On the other hand, several classifiers are applied

**Database**
The database used is the ORL (Olivetti Research Laboratory) faces database [8]. This database contains a set of face images taken between April 1992 and April 1994 at ORL. The database was used in the context of a face recognition project carried out in collaboration with the Speech, Vision and Robotics Group of the Cambridge University Engineering Department.

There are ten different images of each of 40 distinct subjects. For some subjects, the images were taken at different times, varying the lighting, facial expressions (open/closed eyes, smiling / not smiling) and facial details (glasses / no glasses). All the images were taken against a dark homogeneous background with the subjects in an upright, frontal position (with tolerance for some side movement).

**Conditions of the experiments**
Our results have been obtained with the ORL database in the following situation:40 persons, faces 1 to 5 for training, and faces 6 to 10 for testing.
We obtain one model from each training image. During testing each input image is compared against all the models inside the database (40x5=200 in our case) and the model close to the input image (using Mean Square Error criterion) indicates the recognized person.

**Dimensionality reduction using the DCT**
The first experiment consisted on the evaluation of the identification rates as function of the vector dimension. Thus, 200 test (40 persons x 5 test images per person) are performed for each vector dimension (92 different vector dimensions) and the corresponding identification rates are obtained. Figure 7 plots the obtained results for a circular and a rectangular mask. The possible vector lengths for the rectangular mask are $(N')^2 = 1, 4, 9, 16$, etc. Similar results are obtained with circular and rectangular mask, so the latter is chosen because it's easier to implement.

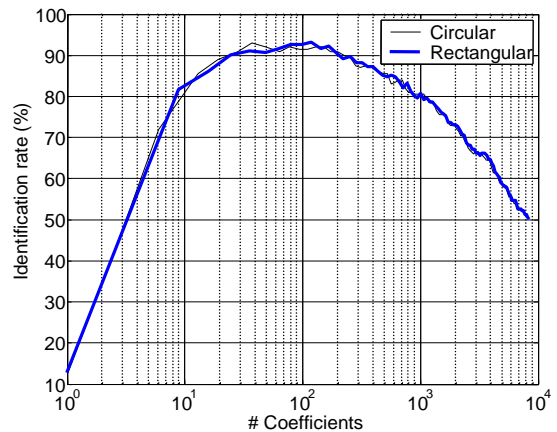

Figure 7 Identification rate vs number of coefficients

The classifier consists on a nearest neighbor classifier using the Mean Square Error (MSE) or the Mean Absolute Difference (MAD) defined as:

$$MSE(\vec{x}, \vec{y}) = \sum_{i=1}^{(N')^2} (x_i - y_i)^2$$

$$MAD(\vec{x}, \vec{y}) = \sum_{i=1}^{(N')^2} |x_i - y_i|$$

Better results are obtained using the MAD criterion. For instance, for a vector dimension of 100 we obtained the following results:
- MSE criterion: identification rate= 91%
- MAD criterion: identification rate = 92.5%

Thus, we have chosen the MAD criterion in our simulations.

Two main conclusions can be obtained from this plot:
- Maximum recognition rates are obtained for a vector dimension around 100. This is the same vector dimension used for the eigenfaces method, where the number of obtained eigenfaces using the approximated method is smaller or equal to the number of classes (persons) by the number of training photographs per person.
- There is a drop in the recognition rates for dimensions higher than 200.

Figure 8 shows the obtained results using the DCT, DFT (implemented with the fast FFT algorithm), and $FFT\{\ln(f(x,y)+0.0001)\}$ in similar conditions than figure 7 (using rectangular mask). Taking into account that the FFT produces complex coefficients and similar recognition rates than DCT, we have chosen the DCT. Thus, the following results will be devoted to this transform. Theoretically the Transform of the logarithm should enable the elimination of lighting artifacts, but the ORL database does not introduce this phenomena, so no advantage is taken in our experiments.

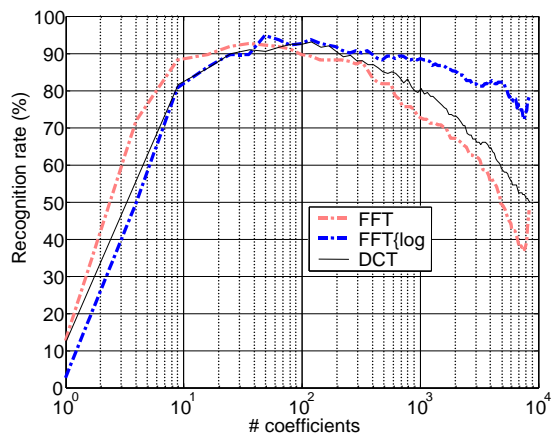

Figure 8, Identification rates vs vector dimension for several transforms.

**Identification rates using a Multi Layer Perceptron (MLP) classifier**

Although the nearest neighbor classifier using the Mean Absolute Difference (MAD) is straightforward to implement, it does not take into account the relevance of different frequency coefficients. On the other hand the performance of the MAD is not good enough when the dimensionality of the vectors is high, which is precisely the case of face recognition. Thus, we use a neural net classifier in order to emphasize those coefficients that contribute more to a good classification (after the training phase the weights assigned to these coefficients would be higher than using the MAD criterion that assigns the same weights to all the coefficients). We use a Multi Layer Perceptron (MLP) trained with a scaled conjugate gradient backpropagation algorithm, that is based on conjugated directions. Figure 9 shows the scheme of the used MLP.

We have set up the following parameters:
- Number or epochs: 15000
- Input neurons: 100
- Hidden layer neurons: 40
- Output neurons: 40 (one per person)
- Performance function: regularized mean square error (MSEREG).

One of the problems that occurs during neural network training is called *overfitting*. The error on the training set is driven to a very small value, but when new data is presented to the network the error is large. The network has memorized the training examples, but it has not learned to generalize to new situations. The adopted solution to the overfitting problem has been the use of regularization. The regularization involves modifying the performance function, which is normally chosen to be the sum of squares of the network errors on the training set. So, this technique helps take the mystery out of how to pick the number of neurons in a network and consistently leads to good networks that are not *overtrained*. The classical Mean Square Error (MSE) implies the computation of: $MSE = \frac{1}{N}\sum_{i=1}^{N}(t_i - a_i)^2$, where t,a are the N dimensional vectors of the test input and the model, respectively. The regularization uses the following measure: $MSEREG = \gamma MSE + (1-\gamma)\frac{1}{n}\sum_{j=1}^{n}w_j^2$.

Thus, it includes one term proportional to the modulus of the weights of the neural net.

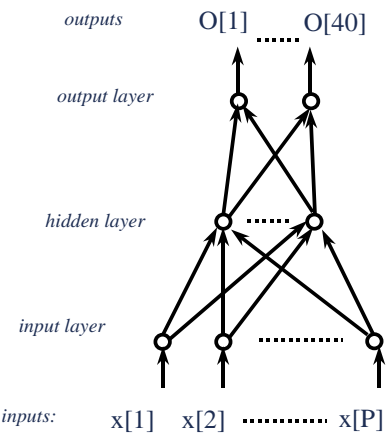

Figure 9 Multi Layer Perceptron

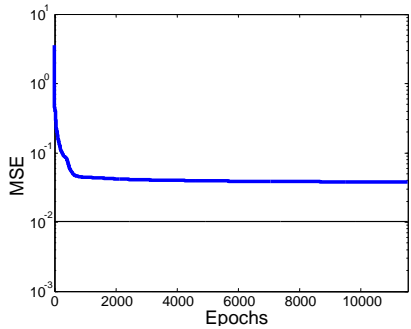

Figure 10: training error vs epochs

The training has been done in the following way: when the input is a genuine face, the output (target of the nnet) is fixed to 1. When the input is an impostor face, the output is fixed to –1. Training and testing sets have been the same than previous sections.

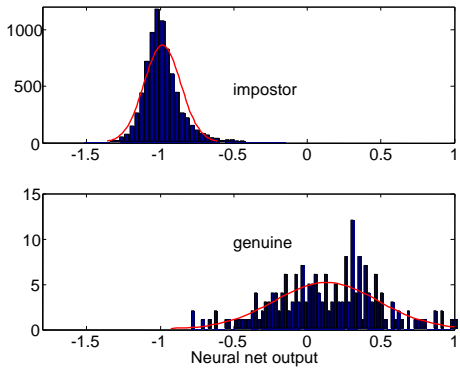

Figure 11 Inter and intra person differences

Figure 10 shows the training error as function of the epochs, and figure 11 the histograms with intra and interpersons distances (genuine and impostor distances repectively) plus a fitted gaussian to the obtained distribution. This plot clearly outperforms our previous results [espinosa] using the classical eigenface approach plus a neural net classifier. It is interesting to observe that there is a preponderance of the negative responses. This is because of the most part of the training vectors are inhibitory. Thus, the nnet tends to learn that "all is inhibitory". Although this, it is possible to discriminate between classes.

**Identification rates using a Radial Basis function network (RBF) and a probabilistic Neural Network (PNN) classifier**

Probabilistic neural networks (PNN) can be used for classification problems. Their design is straightforward and does not depend on training. A PNN is guaranteed to converge to a Bayesian classifier providing it is given enough training data. These networks generalize well.

Given an input, the first layer computes distances from the input vector to the training input vectors, and produces a vector whose elements indicate how close the input is to a training input. The second layer sums these contributions for each class of inputs to produce as its net output a vector of probabilities. Finally, a compete transfer function on the output of the second layer picks the maximum of these probabilities, and produces a 1 for that class and a 0 for the other ones.

If spread is near zero, the network will act as a nearest neighbor classifier. As spread becomes larger, the designed network will take into account several nearby design vectors. We use the following parameters:
- Centers: 200 (each training vector in one center).
- Output neurons: 40 (one per person)

The output of $i^{th}$ Radial Basis neuron is $R_i = radbas(\|\vec{w}_i - \vec{x}\| \times b_i)$, where:

1. $\vec{x}$ is the $p$=100 dimensional input vector
2. $b_i$ is the scalar bias or spread ($\sigma$) of the gaussian
3. $\vec{w}_i$ is the $p$ dimensional weight vector of the Radial Basis neuron $i$.

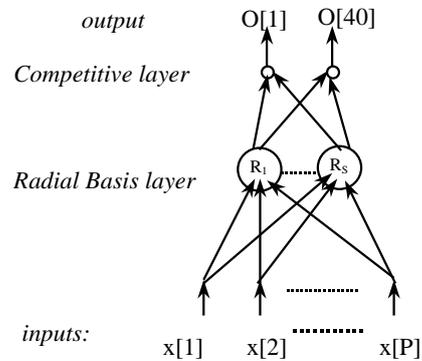

Figure 12 Probabilistic neural network.

Figure 12 shows the architecture of the PNN and figure 13 the identification rates as functions of the spread. Maximum recognition rate is 91%. Although the PNN is faster to train than the MLP, the recognition rates are lower. In order to improve the results and to let the combination with other classifiers, the compete transfer layer is replaced with a layer of pure linear transfer function neurons. The following algorithm is applied till the maximum number of neurons are reached:
1. The network is simulated
2. The input vector with the greatest error is found.
3. A RADBAS neuron is added with weights equal to that vector.
4. The PURELIN layer weights are redesigned to minimize error.

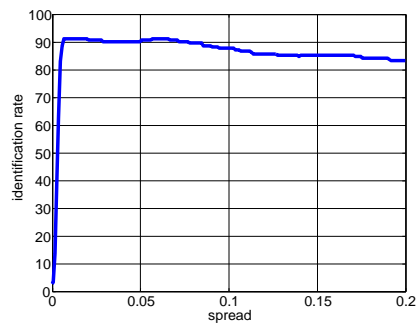

Figure 13. Identification rates vs spread using the PNN

Figure 14 shows the identification rates vs spead using the RBF with 100 centers (radial basis neurons). In this case, the maximum recognition rate is 96% for a spread between 0.8 and 0.9.

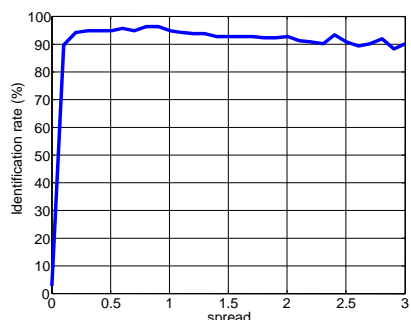

Figure 14. Identification vs spread for the RBF with 100 centers.

**Data fusion**
Several levels of data fusion exist:
- Sensor fusion
- Feature fusion
- Opinion fusion
- Decision fusion

Probably, the most popular is the combination of several classifiers or opinion fusion. This strategy is known as committee machines [10], and it can be considered that it relies on the principle of divide and conquer, where a complex computational task is split into several simpler tasks plus the combination of them. A committee machine is a combination of experts by means of the fusion of knowledge acquired by experts in order to arrive to an overall decision that it is supposedly superior to that attainable by any one of them acting alone. Figure 15 summarizes the scheme.

We have used the mean of the scores obtained with different classifiers [11]

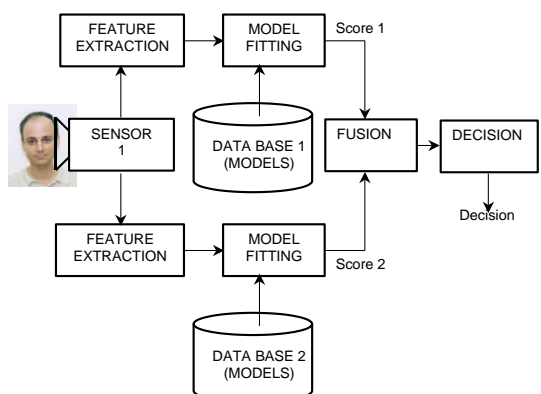

Figure 15 Opinion fusion

**Comparison with the classical eigenfaces approach**
Table 1 compares the results with different feature extraction methods, vector dimensions and distance measure criterion, where NN= Nearest neighbour classifier, RBF= Radial Basis function network, MLP= Multi Layer Perceptron, PNN= Probabilistsic neural network, MAD= Mean Absolute difference, MSE= Mean square error

Table 1. Comparison between different systems

| Feature extraction | Vector dimension | classifier | Identification rate (%) |
|---|---|---|---|
| eigenfaces | 200 | NN (MAD) | 86.5 |
| eigenfaces | 200 | NN (MSE) | 78 |
| eigenfaces | 100 | NN (MAD) | 78.5 |
| eigenfaces | 100 | NN (MSE) | 75.5 |
| DCT | 100 | NN (MAD) | 92.5 |
| DCT | 100 | NN (MSE) | 91 |
| DCT | 100 | MLP | 95 |
| DCT | 100 | RBF | 96 |
| DCT | 100 | PNN | 91 |
| DCT | 100 | RBF+NN (MAD) | 96.5 |

## 5. CONCLUSIONS

In face recognition the image must first be transformed into a low-dimensional coordinate system that preserves the general perceptual quality of the target object's image. This transformation is necessary to address the problem of dimensionality: the raw image data has so many degrees of freedom that it would require millions of examples to directly learn the range of appearances. Typical methods for reducing dimensionality include [12]: KLT (also called Principal Component Analysis or eigenfaces), Ritz Approximation (also called example-based representation), Sparse-filter representations (for example Gabor jets and wavelet transforms), Feature histograms and Independent-Component analysis (ICA). We propose the use of the Discrete Cosine Transform.
The main conclusions are the following:
- The DCT is a good alternative to the classical eigenfaces approach, measured by means of computational burden and recognition accuracy. On the other hand this transform is not data dependent, so it is not necessary to compute any "eigenface".
- There is room for recognition rate improvement in statistical face recognition methods using the classical transform methods applied mainly in image coding.
- We have proven that with a properly training phase, the accuracy of the neural net classifier outperforms the classical approach based on a Nearest Neighbor. We believe that the neural net has a greater generalization capability for recognizing faces not used during training.

**REFERENCES**

[1] R. C. Gonzalez & R. E. Woods "Digital Image Processing" Ed. Addison Wesley 1993

[2] M. Turk M. & A. Pentland, "*Eigenfaces for Recognition*" Journal Cognitive Neuroscience, Vol. 3, nº 1 pp 71-86, Massachusetts Institute of Thecnology.1991.

[3] A. K. Jain "Fundamentals of digital image processing" Ed. Prentice Hall 1989

[4] K. Fukunaga "Statistical Pattern recognition", 2$^{nd}$ edition Ed. Academic Press 1990.

[5] M. Kirby & L. Sirovich "Application of the Karhunen Loeve procedure for the characterization of human faces", IEEE trans. Pattern analysis and machine intelligence , vol. 12 Nº 1 pp.103-108 1990

[6] R. Schalkoff "Pattern recognition statistical, structural and neural approaches" Ed. John Wiley & sons Inc. 1992

[7] Edited by A. Jain, R. Bolle & S. Pankanti, "Biometrics personal identification in networked society" Kluwer Academic Publishers 1999

[8] F. Samaria & A. Harter "Parameterization of a stochastic model for human face identification". 2nd IEEE Workshop on Applications of Computer Vision December 1994, Sarasota (Florida).

[9] V. Espinosa-Duro and M. Faundez-Zanuy, "Face identification by means of a neural net classifier," Proceedings IEEE 33rd Annual 1999 International Carnahan Conference on Security Technology (Cat. No.99CH36303), Madrid 1999, pp. 182-186, doi: 10.1109/CCST.1999.797910

[10] S. Haykin "Neural nets. A comprehensive foundation", 2$^{on}$ edition. Ed. Prentice Hall 1999

[11] J. Kittler, M. Hatef, R. P. W. Duin & J. Matas "On combining classifiers". IEEE Trans. On pattern analysis and machine intelligence, Vol. 20, Nº 3, pp. 226-239, march 1998.

[12] A. Pentland & T. Choudhury "Face recognition for smart environments" IEEE Computer, Vol.33 Nº 2 pp.50-55, February 2000